\documentclass[10pt,twocolumn,letterpaper]{article}

\usepackage{cvpr}
\usepackage{times}
\usepackage{epsfig}
\usepackage{graphicx}
\usepackage{amsmath}
\usepackage{amssymb}

\usepackage[utf8]{inputenc} 
\usepackage[T1]{fontenc}    
\usepackage{hyperref}       
\usepackage{url}            
\usepackage{booktabs}       
\usepackage{amsfonts}       
\usepackage{nicefrac}       
\usepackage{microtype}      
\usepackage{amsmath}
\usepackage{amssymb}

\usepackage[font=footnotesize,labelfont=bf]{caption}

\usepackage{multirow}
\usepackage{tabularx}
\usepackage{dsfont}
\usepackage{url}
\usepackage[tight]{subfigure}
\usepackage{color}
\usepackage{subfigure}
\usepackage{amsthm}
\usepackage[ruled,vlined]{algorithm2e}
\usepackage[export]{adjustbox}
\usepackage{wrapfig}

\cvprfinalcopy 

\def\cvprPaperID{****} 

\begin{document}

\title{Ring loss: Convex Feature Normalization for Face Recognition}


\author{Yutong Zheng, Dipan K. Pal and Marios Savvides \\
  Department of Electrical and Computer Engineering\\
  Carnegie Mellon University\\
{\tt\small \{yutongzh, dipanp, marioss\}@andrew.cmu.edu}
}

\maketitle

\begin{abstract}
We motivate and present Ring loss, a simple and elegant feature normalization approach for deep networks designed to augment standard loss functions such as Softmax. We argue that deep feature normalization is an important aspect of supervised classification problems where we require the model to represent each class in a multi-class problem equally well. The direct approach to feature normalization through the hard normalization operation results in a non-convex formulation. Instead, Ring loss applies soft normalization, where it gradually learns to constrain the norm to the scaled unit circle while preserving convexity leading to more robust features. We apply Ring loss to large-scale face recognition problems and present results on LFW, the challenging protocols of IJB-A Janus, Janus CS3 (a superset of IJB-A Janus), Celebrity Frontal-Profile (CFP) and MegaFace with 1 million distractors. Ring loss outperforms strong baselines, matches state-of-the-art performance on IJB-A Janus and outperforms all other results on the challenging Janus CS3 thereby achieving state-of-the-art. We also outperform strong baselines in handling extremely low resolution face matching.
\end{abstract}


\section{Introduction}

Deep learning has demonstrated impressive performance on a variety of tasks. Arguably the most important task, that of supervised classification, has led to many advancements. Notably, the use of deeper structures  \cite{simonyan2014very, szegedy2015going, he2016deep} and more powerful loss functions \cite{hadsell2006dimensionality, schroff2015facenet, wen2016discriminative, tadmor2016learning, liu2016large} have resulted in far more robust feature representations.  There has also been more attention on obtaining better-behaved gradients through normalization of batches or weights \cite{ioffe2015batch, ba2016layer, salimans2016weight}. 


One of the most important practical applications of deep networks with supervised classification is face recognition. Robust face recognition poses a huge challenge in the form of very large number of classes with relatively few samples per class for training with significant nuisance transformations. A good understanding of the challenges in this task results in a better understanding of the core problems in supervised classification, and in general representation learning. However, despite the impressive attention on face recognition tasks over the past few years, there are still many gaps towards such an understanding. Notably, the need and practice of feature normalization. Normalization of features has recently been discovered to provide significant improvement in performance which implicitly results in a cosine embedding \cite{ranjan2017l2, Wang2017NormFace}. However, direct normalization in deep networks explored in these works results in a non-convex formulation resulting in local minima generated by the loss function itself. It is important to preserve convexity in loss functions for more effective minimization of the loss given that the network optimization itself is non-convex.  In a separate thrust of work, cosine similarity has also been very recently explored for supervised classification \cite{liu2017learning, chunjie2017cosine}. Nonetheless, a concrete justification and principled motivation for the \textit{need} for normalizing the features itself is also lacking.


\begin{figure}
    \begin{center}
        \subfigure[Features trained using Softmax]{%
        \centering
            \includegraphics[width=0.45\columnwidth,height=0.4\columnwidth,valign=m]{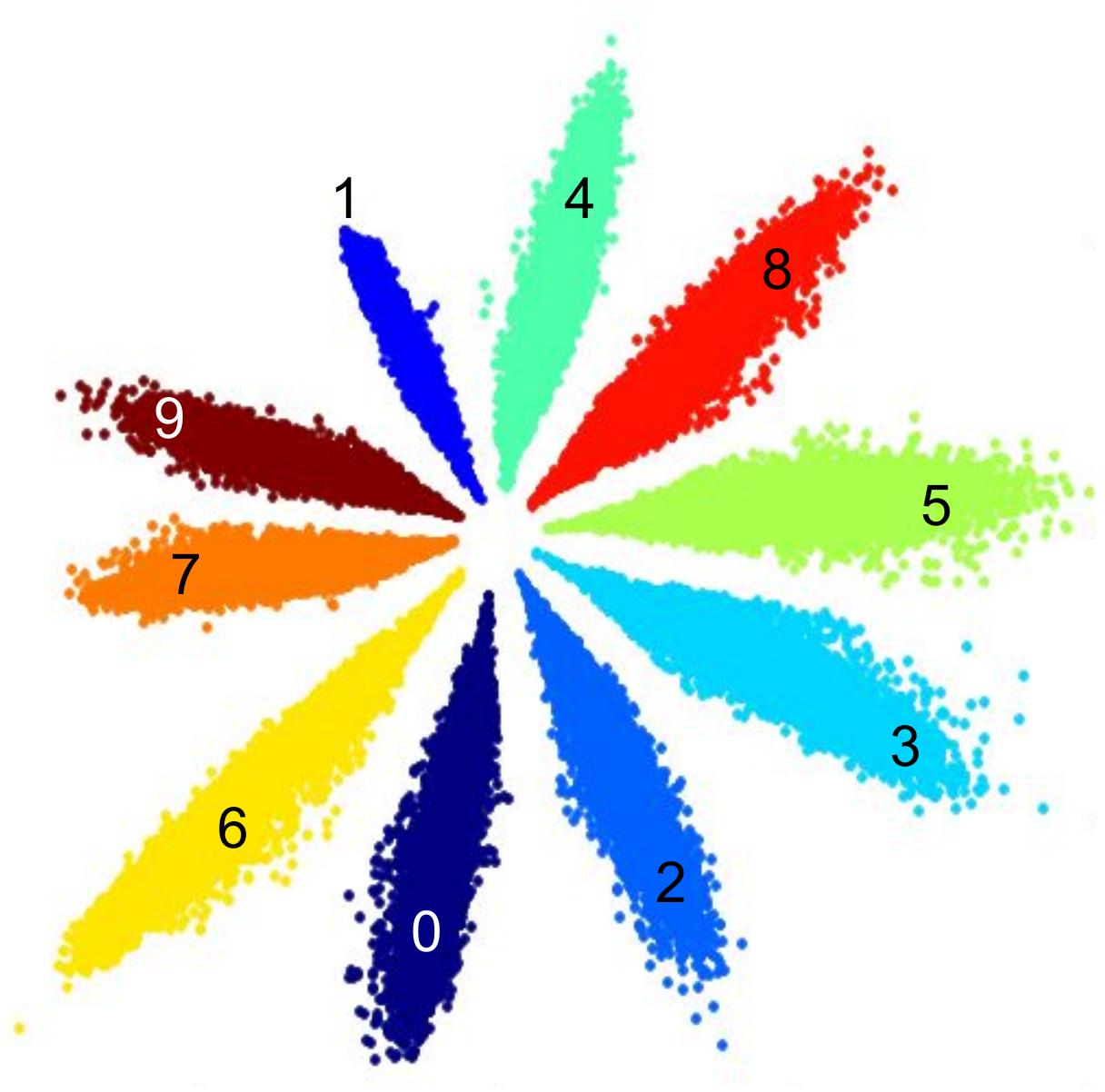}\label{fig_softmax}
        }%
         \subfigure[Features trained using Ring loss]{%
        \centering
        \includegraphics[width=0.5\columnwidth,height=0.4\columnwidth,valign=m]{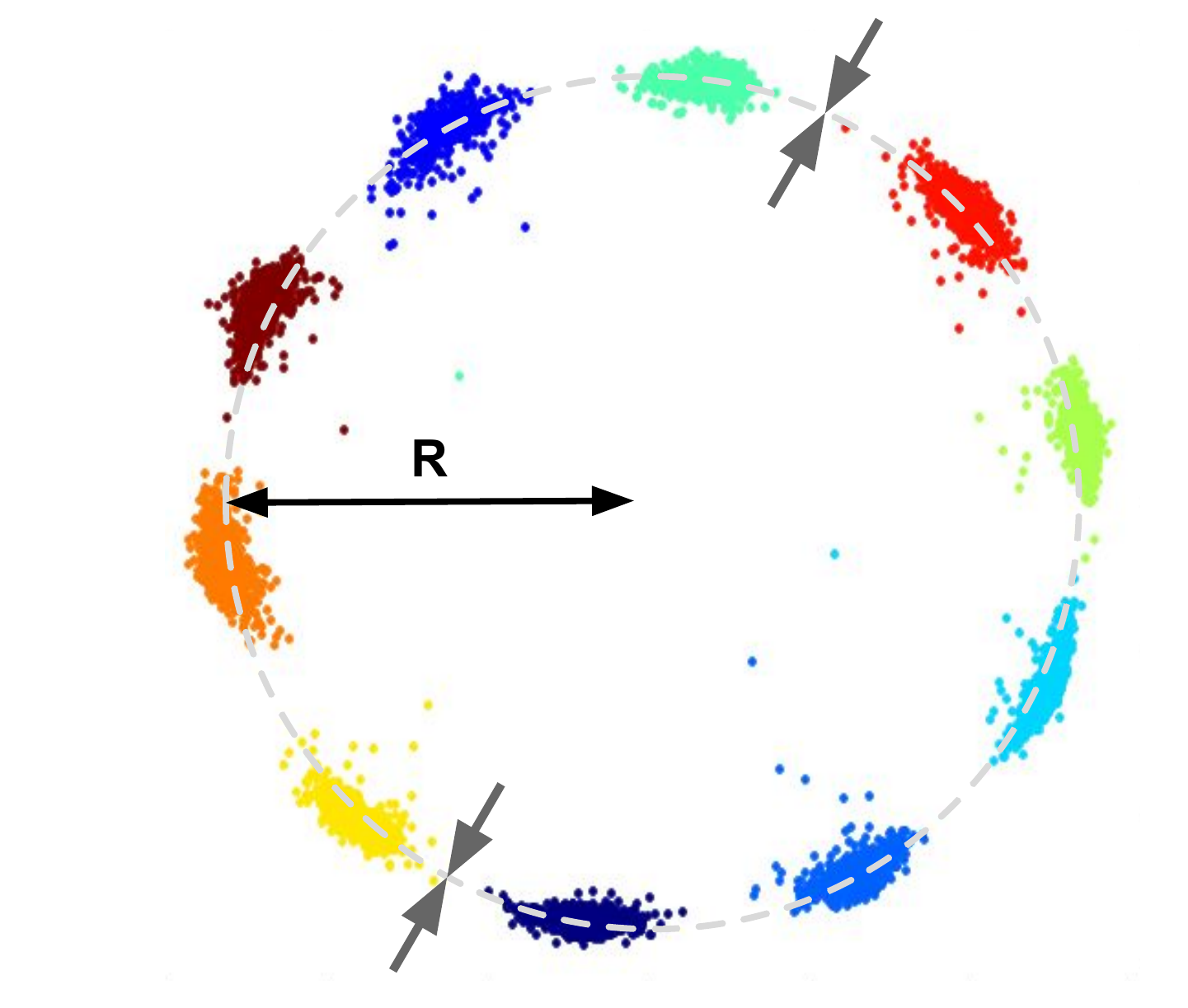}\label{fig_ring_loss}
        }
    \end{center}
    \vspace{-0.5cm}
\caption{Sample MNIST features trained using (a) Softmax and (b) Ring loss on top of Softmax. Ring loss uses a convex norm constraint to gradually enforce normalization of features to a learned norm value $R$. This results in features of equal length while mitigating classification margin imbalance between classes. Softmax achieves 98.97 \% accuracy on MNIST, whereas Ring loss achieves 99.34 \% demonstrating the superior performance of the network learned normalized features. } 
\label{fig_sample_visualization}
\vspace{-0.5cm}
\end{figure}

\begin{figure*}

    \begin{center}
        \subfigure[A 2-class 2-sample case]{%
        \centering
            \includegraphics[width=0.7\columnwidth,valign=m]{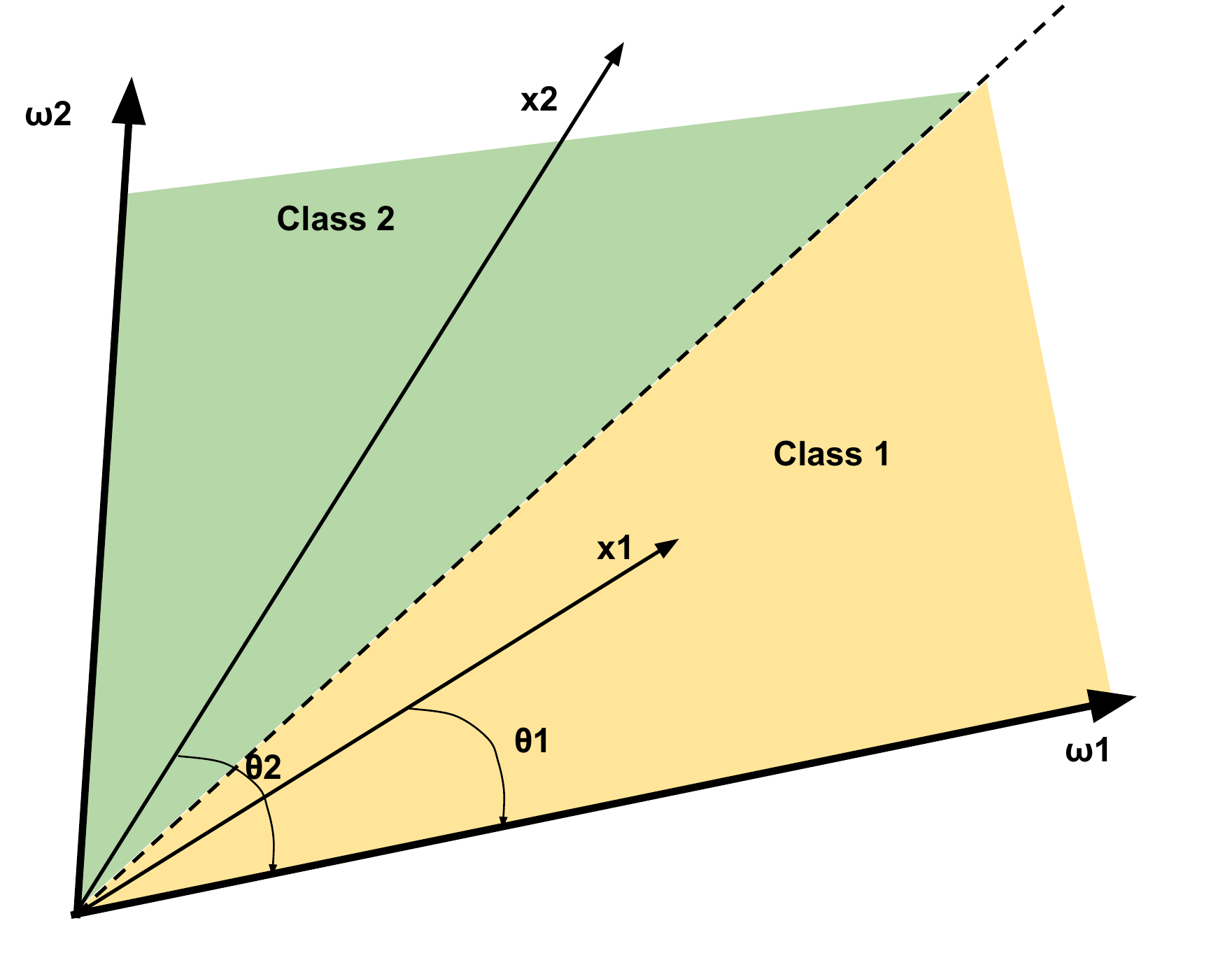}\label{fig_class_margin_setting}
        }%
         \subfigure[Classification margin]{%
        \centering
        \includegraphics[width=0.9\columnwidth,valign=m]{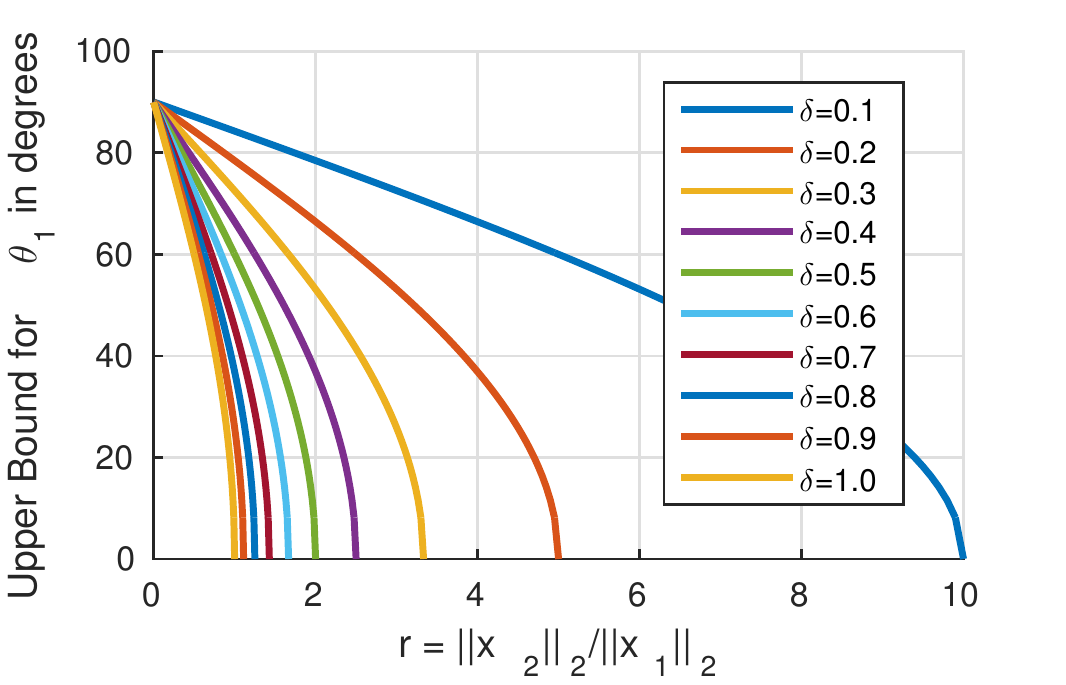}\label{fig_class_margin}
        }
    \end{center}
    \vspace{-0.7cm}
\caption{  (a) A simple case of binary classification.  The shaded regions (yellow, green)  denote the classification margin (for class 1 and 2).  (b)  Angular classification margin for $\theta_1$ for different $\delta=\cos \theta_2$. }
\label{fig_class_margin_visualization}
\vspace{-0.5cm}
\end{figure*}

\textbf{Contributions.} In this work, we propose Ring loss, a simple and elegant approach to normalize all sample features through a convex augmentation of the primary loss function (such as Softmax). The value of the target norm is also learnt during training. Thus, the only hyperparameter in Ring  loss is the loss weight w.r.t to the primary loss function. We provide an analytical justification illustrating the benefits of feature normalization and thereby cosine feature embeddings. Feature matching during testing in face recognition is typically done through cosine distance creating a gap between testing and training protocols which do not utilize normalization. The incorporation of Ring loss during training eliminates this gap. Ring loss is differentiable that allows for seamless and simple integration into deep architectures trained using gradient based methods. We find that Ring loss provides consistent improvements over a large range of its hyperparameter when compared to other baselines in normalization and indeed other losses proposed for face recognition in general. Interestingly, we also find that Ring loss helps in being robust to lower resolutions through the norm constraint.




\section{Ring loss: Convex Feature Normalization}

\subsection{Intuition and Motivation. }

\begin{figure*}[t]
\centering
\begin{tabular}{ccc}
\subfigure[$\lambda=0$  (Pure Softmax) ]{\includegraphics[scale=.3]{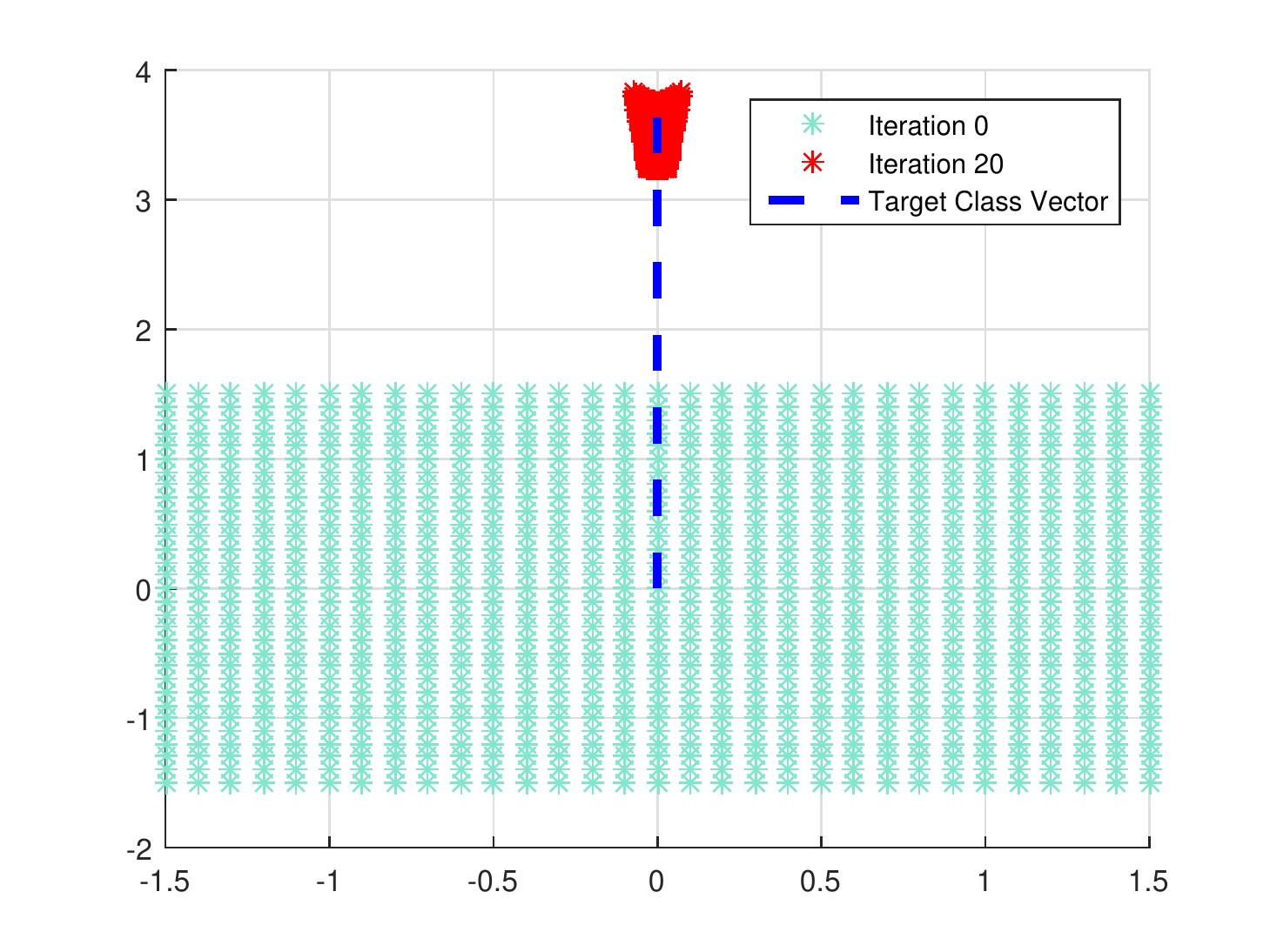}} & 
\subfigure[$\lambda=1$ ]{\includegraphics[scale=.3]{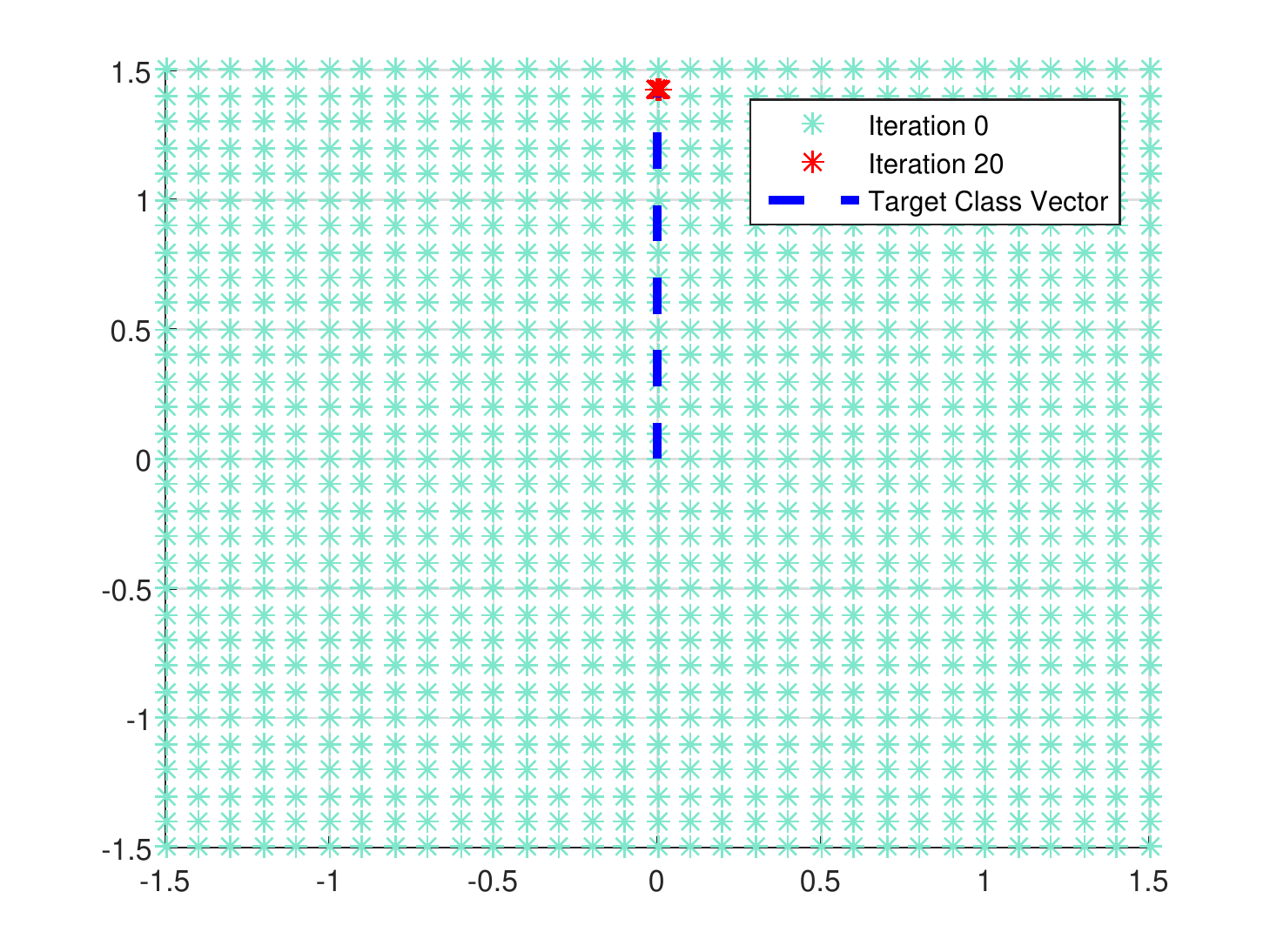}} &
\subfigure[$\lambda=10$ ]{\includegraphics[scale=.3]{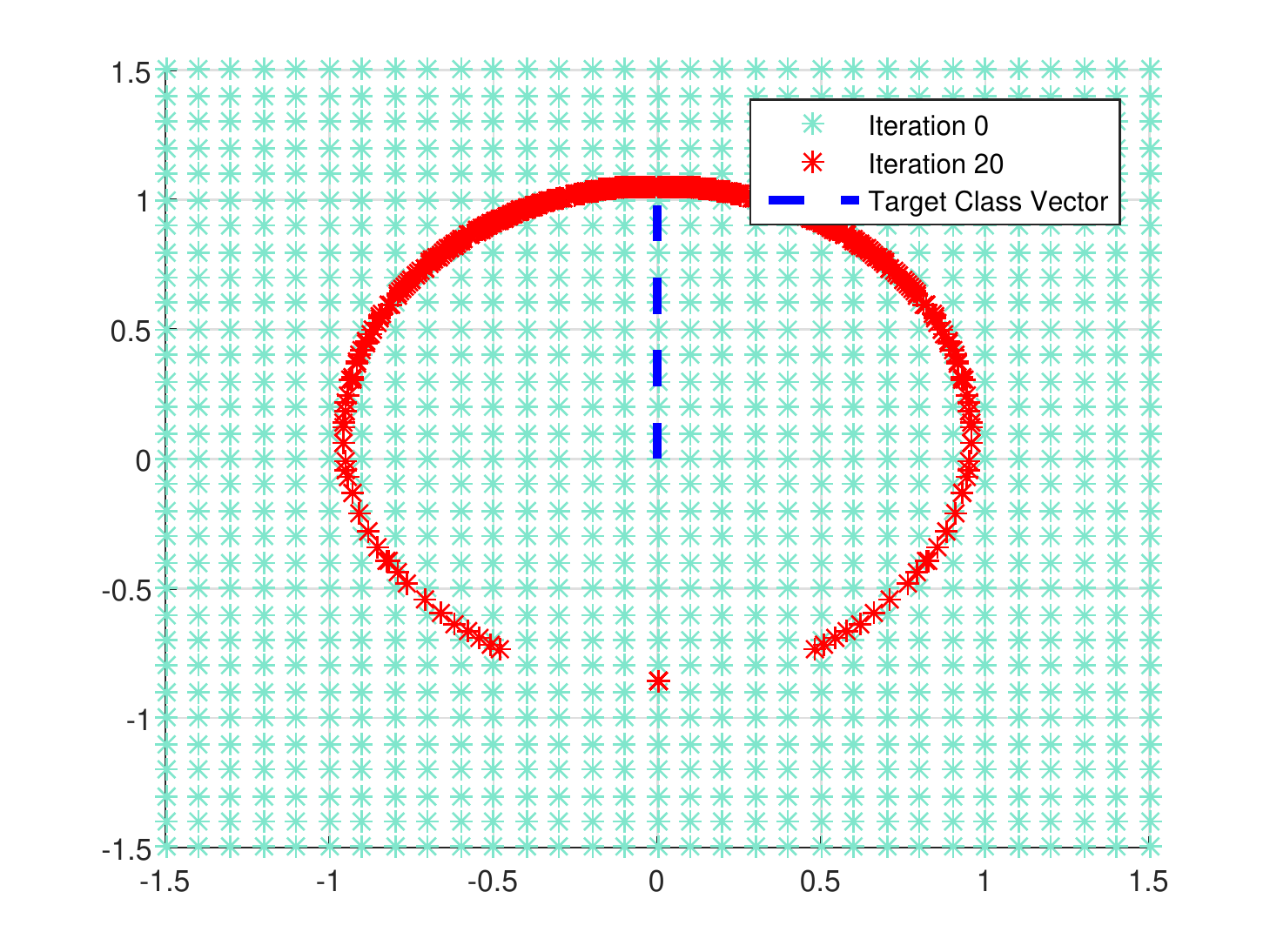}}

\end{tabular}
\vspace{-0.5cm}
\caption{\textbf{Ring loss Visualizations:} (a), (b) and (c) show the final convergence of the samples (for varying $\lambda$). The blue-green dots are the samples before the gradient update and the red dots are the same samples after the update. The dotted blue vector is the target class direction. $\lambda=0$ fails to converge and does not constrain the norm whereas $\lambda=10$ takes very small steps towards Softmax gradients. A good balance is achieved at  $\lambda=1$. In our large scale experiments, a large range of $\lambda$ achieves this balance. }
\label{fig_ring_loss_visual}
\vspace{-0.5cm}
\end{figure*}

 
 There have been recent studies on the use of norm constraints right before the Softmax loss \cite{Wang2017NormFace, ranjan2017l2}. However, the formulations investigated are non-convex in the feature representations leading to difficulties in optimization. Further, there is a need for better understanding of the benefits of normalization itself. Wang \emph{et.al.} \cite{Wang2017NormFace} argue that the `radial' nature of the Softmax features is not a useful property, thereby cosine similarity should be preferred leading to normalized features. A concrete reason was, however, not provided. Ranjan \emph{et.al.} \cite{ranjan2017l2} show that the Softmax loss encodes the quality of the data (images) into the norm thereby deviating from the ultimate objective of learning a good representation purely for classification.\footnote{We in fact, find in our pilot study that the Softmax features also encode the `difficulty' of the class.} Therefore for better classification, normalization forces the network to be invariant to such details. This is certainly not the entire story and in fact overlooks some key properties of feature normalization. We now motivate Ring loss with three arguments. 1) We show that the norm constraint is beneficial to maintain a balance between the angular classification margins of multiple classes. 2) It removes the disconnect between training and testing metrics. 3) It minimizes test errors due to angular variation due to low norm features.   
 


\textbf{The Angular Classification Margin Imbalance. } Consider a binary classification task with two feature vectors $x_1$ and $x_2$ from class 1 and 2 respectively, extracted using some model (possibly a deep network). Let the classification weight vector for class 1 and 2 be $w_1, w_2$ respectively (potentially Softmax). An example arrangement is shown in Fig.~\ref{fig_class_margin_setting}. Then in general, in order for the class 1 vector $w_1$ to pick $x_1$ \textit{and not} $x_2$ for correct classification, we require $w_1^T x_1 > w_1^Tx_2 \Rightarrow  \|x_1\|_2 \cos \theta_1  >  \|x_2\|_2\cos \theta_2$\footnote{Although, it is more common to in turn investigate competition between two weight vectors to classify a single sample, we find that this alternate perspective provide some novel and interesting insights.}. Here, $\theta_1$ and $\theta_2$ are the angles between the weight vector $w_1$ (class 1 vector only) and $x_1$, $x_2$ respectively\footnote{Note that this reasoning is applicable to any loss function trying to enforce this inequality in some form.}.  We call the feasible set (range for $\theta_1$) for this inequality to hold as the \textit{angular classification margin}. Note that it is also a function of $\theta_2$.  Setting $\frac{\|x_2\|_2}{\|x_1\|_2} = r$, we observe $r>0$ and that for correct classification, we need $\cos \theta_1 > r \cos \theta_2 \Rightarrow \theta_1 < \cos ^{-1}( r \cos \theta_2)$ since $\cos \theta$ is a decreasing function between $[-1, 1]$ for $\theta \in [0, \pi]$. This inequality needs to hold true for any $\theta_2$. Fixing $\cos \theta_2 = \delta$, we have $\theta_1 < \cos ^{-1} (r\delta)$. From the domain constraints of $\cos^{-1}$, we have $-1 \leq r\delta \leq 1 \Rightarrow \frac{-1}{\delta}  \leq r \leq \frac{1}{\delta }$. Combining this inequality with $r>0$, we have $ 0 < r \leq \frac{1}{|\delta|} \Rightarrow  \|x_2\|_2  \leq \frac{1}{\delta } \|x_1\|_2 ~~\forall  \delta \in (0 ~1]$. For our purposes it suffices to only look at the case $\delta > 0$ since the $\delta<0$ doesn't change the inequality $-1 \leq r\delta \leq 1 $ and is more interesting.


\textbf{Discussion on the angular classification margin. }  We plot the upper bound on $\theta_1$ (\emph{i.e.} $\cos ^{-1}( r \cos \theta_2)$) for a range of $\delta$ ([0.1, 1]) and the corresponding range of $r$. Fig.~\ref{fig_class_margin} showcases the plot. Consider $\delta=0.1$ which implies that the sample $x_2$ has a large angular distance from $w_1$ (about $85^\circ$). This case is favorable in general since one would expect a lower probability of $x_2$ being classified to class 1. However, we see that as $r$ increases (difference in norm of $x_1, x_2$), the classification margin for $x_1$ decreases from $90^\circ$ to eventually $0^\circ$. In other terms, as the norm of $x_2$ increases w.r.t $x_1$, the angular margin for $x_1$ to be classified correctly while rejecting $x_2$ by $w_1$, decreases. The difference in norm ($r>1$) therefore will have an adverse effect during training by effectively enforcing smaller angular classification margins for classes with smaller norm samples. This also leads to lop-sided classification margins for multiple classes due to the difference in class norms as can be seen in Fig.~\ref{fig_softmax}. This effect is only magnified as $\delta$ increases (or the sample $x_2$ comes closer to $w_1$). Fig.~\ref{fig_class_margin} shows that the angular classification margin decreases much more rapidly as $\delta$ increases. However, $r<1$ leads to a larger margin and seems to be beneficial for classifying class 1 (as compared to $r>1$). One might argue that this suggests that the $r<1$ should be enforced for better performance. However, note that the same reasoning applies correspondingly to class 2, where we want to classify $x_2$ to $w_2$ while rejecting $x_1$. This creates a trade off between performance on class 1 versus class 2 based on $r$ which also directly scales to multi-class problems. In typical recognition applications such as face recognition, this is not desirable. Ideally, we would want to represent all classes equally well. Setting $r=1$ or constraining the norms of the samples from both classes to be the same ensures this.

\textbf{Effects of Softmax on the norm of MNIST features.} We qualitatively observe the effects of vanilla Softmax on the norm of the features (and thereby classification margin) on MNIST in Fig.~\ref{fig_softmax}. We see that digits 3, 6 and 8 have large norm features which are typically the classes that are harder to distinguish between. Therefore, we observe $r<1$ for these three `difficult' classes (w.r.t to the other `easier' classes) thereby providing a larger angular classification margin to the three classes. On the other hand, digits 1, 9 and 7 have lower norm corresponding to $r>1$ w.r.t to the other classes, since the model can afford to decrease the margin for these `easy' classes as a trade off. We also observe that arguably most easily distinguishable class, digit 1, has the lowest norm thereby the highest $r$. On the other hand, Fig.~\ref{fig_ring_loss} showcases the features learned using Softmax augmented with our proposed Ring loss, which forces the network to learn feature normalization through a convex formulation thereby mitigating this imbalance in angular classification margins. 

\textbf{ Regularizing Softmax loss with the norm constraint.}
The ideal training scenario for a system testing under the cosine metric would be where all features pointing in the same \textit{direction} have the same loss. However, this is not true for the most commonly used loss function, Softmax and its variants (FC layer combined with the softmax function and the cross-entropy loss). Assuming that the weights are normalized, i.e. $\|w_k\|=1$, the Softmax loss for feature vector $\mathcal{F}(\mathbf{x}_i)$ can be expressed as (for the correct class $y_i$):
\begin{align}
L_{SM} &= -\log{
 \frac{
  \exp{w_k \mathcal{F}(\mathbf{x}_i)}}
  {
  \sum_{{k'}=1}^K 
   {\exp{w_{k'} \mathcal{F}(\mathbf{x}_i)}}
  }F
 }\\
 &= -\log{
 \frac{
  \exp{\|\mathcal{F}(\mathbf{x}_i)\|\cos{\theta_{ki}}}}
  {
  \sum_{{k'}=1}^K 
   {\exp{\|\mathcal{F}(\mathbf{x}_i)\|\cos{\theta_{k'i}}}}
  }
 }
\end{align}

Clearly, despite having the same direction, two features with different norms have different losses. From this perspective, the straightforward solution to regularize the loss and remove the influence of the norm is to normalize the features before Softmax as explored in $l_2$-constrained Softmax  \cite{ranjan2017l2}. However, this approach is effectively a projection method, i.e. it calculates the loss as if the features are normalized to the same scale, while the actual \textit{network} does not \textit{learn} to normalize features.

\begin{figure}
    \begin{center}
        \subfigure[Features for training set]{%
        \centering
            \includegraphics[width=0.49\columnwidth,height=0.4\columnwidth,valign=m]{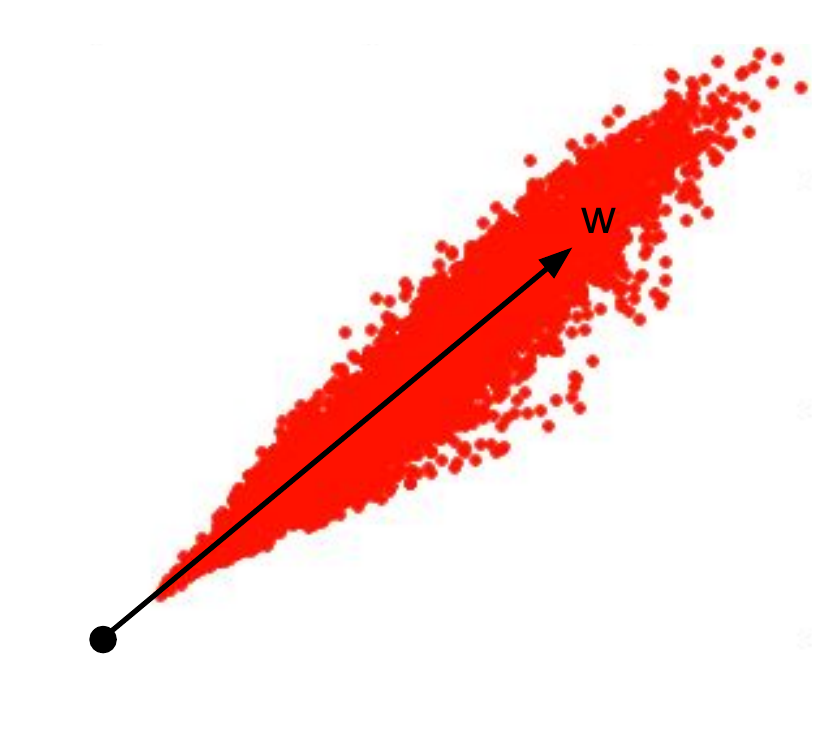}\label{fig_softmax_train}
        }%
         \subfigure[Features for testing set]{%
        \centering
        \includegraphics[width=0.49\columnwidth,height=0.4\columnwidth,valign=m]{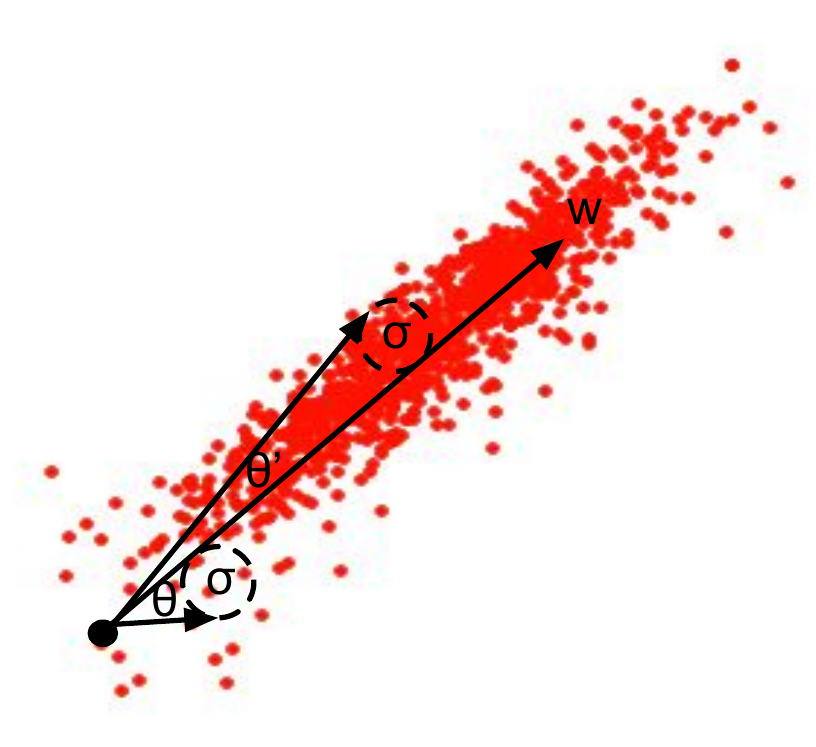}\label{fig_softmax_test}
        }
    \end{center}
    \vspace{-0.5cm}
\caption{MNIST features for digit 8 trained using vanilla Softmax loss. } 
\label{fig_train_test}
\vspace{-0.7cm}
\end{figure}

\textbf{ The need for features normalization in feature space.} As an illustration, consider the training and testing set features trained by vanilla Softmax, of the digit 8 from MNIST in Fig.~\ref{fig_train_test}. Fig.~\ref{fig_softmax_train} shows that at the end of training, the features are well behaved with a large variation in the norm of the features with a few samples with low norm. However, Fig.~\ref{fig_softmax_test} shows that that the features for the test samples are much more erratic. There is a similar variation in norm but now most of the low norm features have huge variation in \textit{angle}. Indeed, variation in samples for lower norm features translates to a larger variation in angle than the same for higher norm samples features. This translates to higher errors in classification under the cosine metric (as is common in face recognition). This is yet another motivation to normalize features during training. Forcing the network to \textit{learn} to normalize the features helps to mitigate this problem during test wherein the network learns to work in the normalized feature space. A related motivation to feature normalization was proposed by Ranjan \emph{et.al.} \cite{ranjan2017l2} wherein it was argued that low resolution of an image results in a low norm feature leading to test errors. Their solution to project (not implicitly learn) the feature to the scaled unit hypersphere was also aimed at handling low resolution. We find in our large scale experiment with low resolution images (see Exp. 6 Fig.~\ref{fig_scale2}) that soft normalization by Ring loss achieves better results. In fact hard projection method by $l_2$-constrained Softmax \cite{ranjan2017l2} performs worse than Softmax for a downsampling factor of 64.



\textbf{ Incorporating the norm constraint as a convex problem.} Identifying the need to normalize the sample features from the network, we now formulate the problem. We define $L_S$ as the primary loss function (for instance Softmax loss). Assuming that $\mathcal{F}$ provides deep features for a sample $x$ as $\mathcal{F}(x)$, we would like to minimize the loss subject to the normalization constraint as follows,
\begin{align}
\min L_S(\mathcal{F}(x)) ~~s.t. ~~ \|\mathcal{F}(x)\|_2=R \label{eq_1}
\end{align}
Here, $R$ is the scale constant that we would like the features to be normalized to. This is the exact formulation recently studied and implemented by \cite{ranjan2017l2, Wang2017NormFace}. Note that this problem is non-convex in $\mathcal{F}(x)$ since the set of feasible solutions is itself non-convex due to the norm equality constraint. Approaches which use standard SGD while ignoring this critical point would not be providing feasible solutions to this problem thereby, the network $\mathcal{F}$ would not learn to output normalized features. Indeed, the features obtained using this straightforward approach are not normalized as was found in Fig. 3b in \cite{ranjan2017l2} compared to our approach (Fig.~\ref{fig_ring_loss}). One naive approach to get around this problem would be to relax the norm equality constraint to an inequality. This objective will now be convex, however it does not necessarily enforce equal norm features. In order to incorporate the formulation as a convex constraint, the following form is directly useful as we find below.




\subsection{Ring loss}

\textbf{Ring loss Definition. } Ring loss $L_R$ is defined as
\begin{align} 
L_R = \frac{\lambda}{2m} \sum_{i=1}^m (\|\mathcal{F}(\mathbf{x}_i)\|_2  -  R)^2
\end{align}
where $\mathcal{F}(\mathbf{x}_i)$ is the deep network feature for the sample $\mathbf{x}_i$.  Here, $R$ is the target norm value which is also learned and $\lambda$ is the loss weight enforcing a trade-off between the primary loss function. $m$ is the batch-size. The square on the norm difference helps the network to take larger steps when the norm of a sample is too far off from $R$ leading to faster convergence. The corresponding gradients are as follows.
\begin{align}
\frac{\partial L_R}{\partial R } &= -\frac{\lambda}{m}\sum_{i=1}^m (\|\mathcal{F}(\mathbf{x}_i)\|_2  -  R) \\ \frac{\partial L_R}{\partial \mathcal{F}(\mathbf{x}_i) }&=  \frac{\lambda}{m}  \left(1  -  \frac{R}{\|\mathcal{F}(\mathbf{x}_i)\|_2}  \right) \mathcal{F}(\mathbf{x}_i) 
\end{align}

\begin{figure}
    \begin{center}
        \subfigure{%
        \centering
        \includegraphics[width=0.8\columnwidth,height=0.35\columnwidth,valign=m]{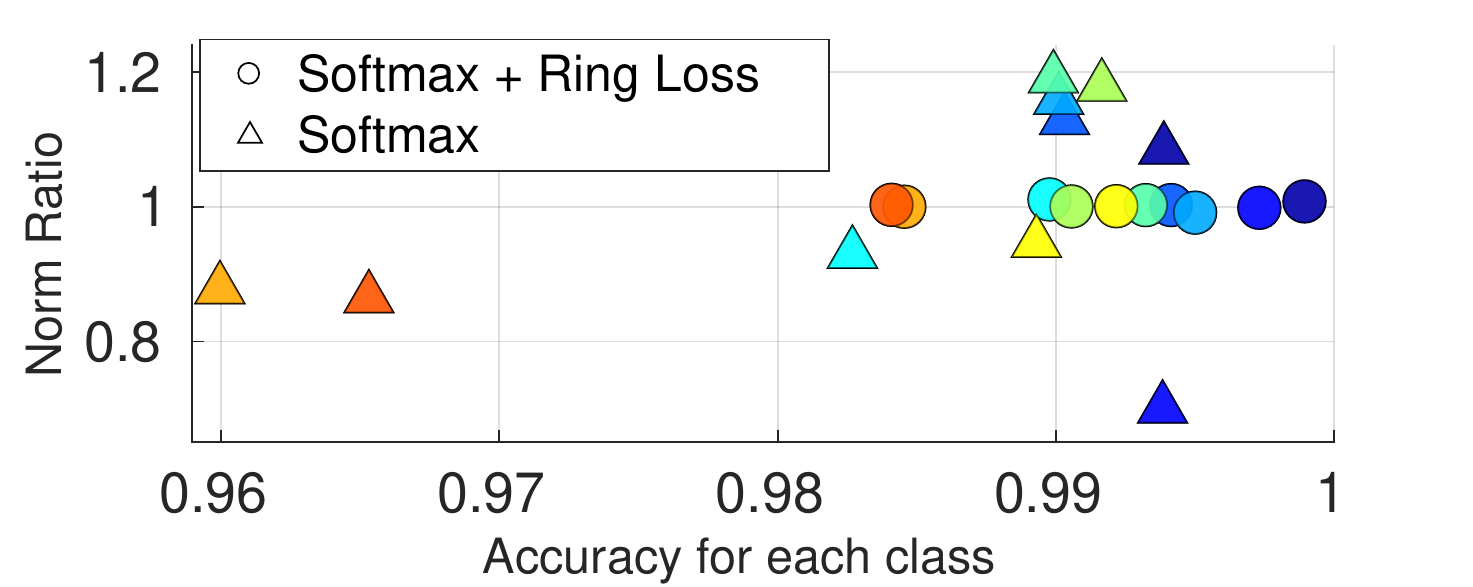}
        }
        \hspace{-0.85cm}
        \subfigure{%
        \centering
        \includegraphics[width=0.15\columnwidth,height=0.30\columnwidth,valign=m]{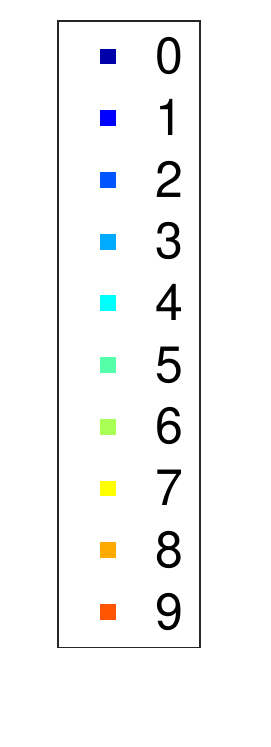}
        }
    \end{center}
    \vspace{-0.5cm}
\caption{Ring loss improves MNIST testing accuracy across all classes by reducing inter-class norm variance. Norm Ratio is the ratio between average class norm and average norm of all features.} 
\label{fig_accuracy}
\vspace{-0.7cm}
\end{figure}

Ring loss ($L_R$) can be used along with any other loss function such as Softmax or large-margin Softmax \cite{liu2017sphereface}. The loss encourages norm of samples being value $R$ (a learned parameter) rather than explicit enforcing through a hard normalization operation. This approach provides informed gradients towards a better minimum which helps the network to satisfy the normalization constraint. The network therefore, learns to normalize the features using \textit{model weights themselves} (rather than needing an explicit non-convex normalization operation as in \cite{ranjan2017l2}, or batch normalization \cite{ioffe2015batch}). In contrast and in connection, batch normalization \cite{ioffe2015batch} enforces the scaled normal distribution for each element in the feature independently. This does not constrain the overall norm of the feature to be equal across all samples and neither addresses the class imbalance problem. As shown in Fig. \ref{fig_accuracy}, Ring loss stabilizes the feature norm across all classes, and, in turn, rectifies the classification imbalance for Softmax to perform better overall. 

\textbf{Ring loss Convergence Visualizations.} To illustrate the effect of the Softmax loss augmented with the enforced soft-normalization, we conduct some analytical simulations. We generate a 2D mesh of points from $( -1.5, 1.5 )$ in x,y-axis. We then compute the gradients of Ring loss ($R=1$) assuming the dottef blue vertical line (see Fig.~\ref{fig_ring_loss_visual}) as the target class and update each point with a fixed step size for 20 steps. We run the simulation for $\lambda = \{ 0, 1, 10 \}$. Note that $\lambda=0$ represents pure Softmax. Fig.~\ref{fig_ring_loss_visual} depicts the results of these simulations.  Sub-figures (a), (b) and (c)  in Fig.~\ref{fig_ring_loss_visual} show the initial points on the mesh grid (light green) and the final updated points (red). For pure Softmax ($\lambda=0$), we see that the updates increases norm of the samples and moreover they do not converge. For a reasonable loss weight of $\lambda=1$, Ring loss gradients can help the updated points converge much faster in the same number of iterations. For heavily weighted Ring loss with $\lambda=10$, we see that the gradients force the samples to a unit norm since $R$ was set to 1 while overpowering Softmax gradients. These figures suggest that there exists a trade off enforced by $\lambda$ between the Softmax loss $L_S$ and the normalization loss. We observe similar trade-offs in our experiments.\\

\section{Experimental Validation}


We benchmark Ring loss on large scale face recognition tasks while augmenting two loss functions. The first one is the ubiquitous Softmax, and the second being a successful variant of Large-margin Softmax \cite{liu2016large} called SphereFace \cite{liu2017sphereface}.  We present results on five large-scale benchmarks of LFW \cite{LFWTech}, IARPA Janus Benchmark IJB-A   \cite{klare2015pushing}, Janus Challenge Set 3 (CS3) dataset (which is a super set of the IJB-A Janus dataset), Celebrities Frontal-Profile (CFP) \cite{sengupta2016frontal} and finally the MegaFace dataset \cite{kemelmacher2016megaface}. We also present results of Ring loss augmented Softmax features on low resolution images from Janus CS3 to showcase resolution robust face matching.

\begin{figure}
        \centering
            \includegraphics[width=0.9\columnwidth,valign=m]{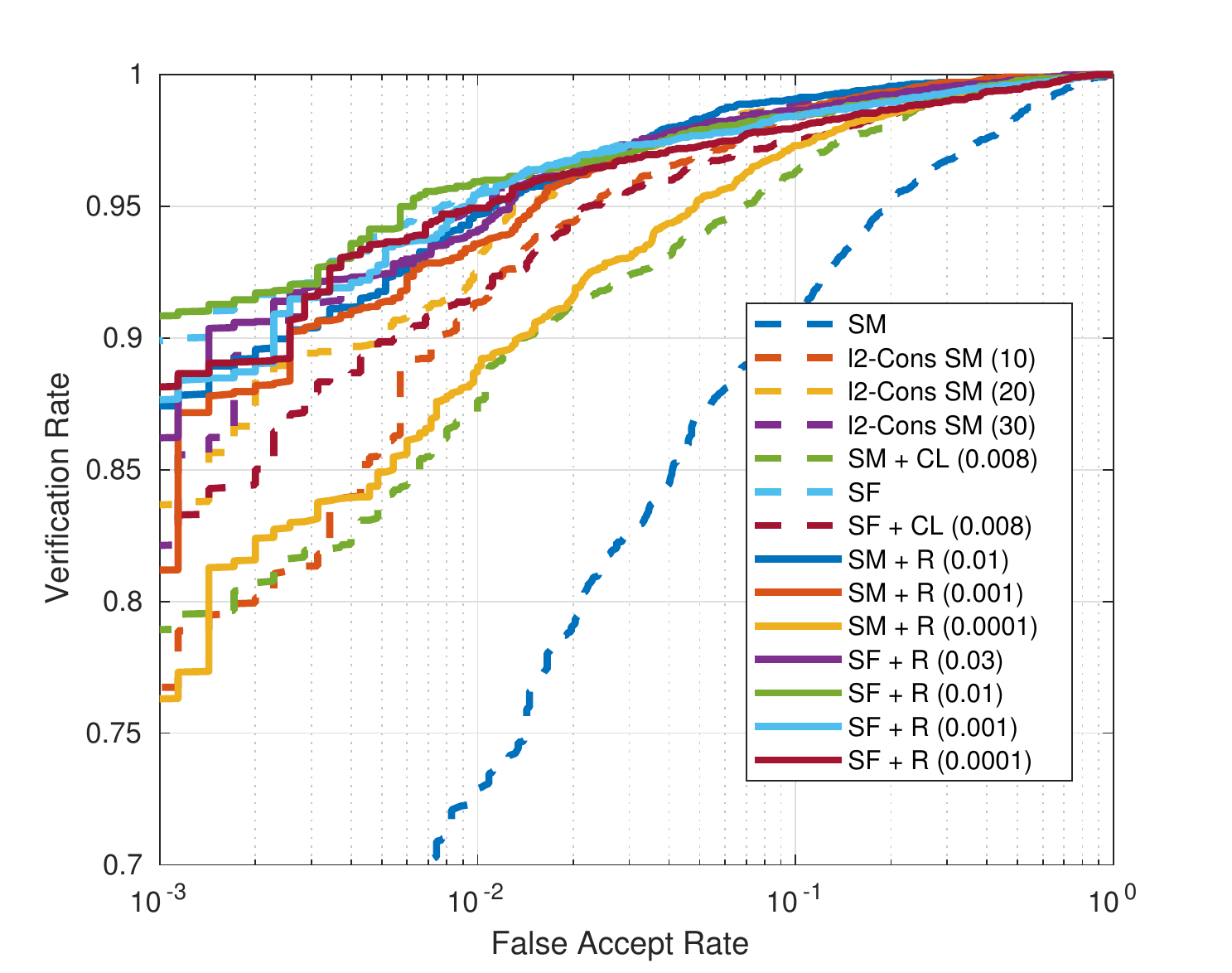}
        
    \vspace{-0.2cm}
\caption{ ROC curves on the CFP Frontal vs. Profile verification protocol. For all Figures and Tables, SM denotes Softmax, SF denotes SphereFace \cite{liu2017sphereface},l2-Cons SM denotes \cite{ranjan2017l2}, + CL denotes Center Loss augmentation \cite{wen2016discriminative} and finally + R denotes Ring loss augmentation. Numbers in bracket denote value of hyperparameter (loss weight), \emph{i.e.} $\alpha$ for \cite{ranjan2017l2}, $\lambda$ for Center loss and Ring loss.}
\label{fig_cfp}
\vspace{-0.5cm}
\end{figure}


\textbf{Implementation Details.}
For all the experiments in this paper, we usethe ResNet 64 (Res64) layer architecture from Liu \emph{et. al.} \cite{liu2017sphereface}. For Center loss, we utilized the code repository online and used the best hyperparameter setting reported\footnote{see \url{https://github.com/ydwen/caffe-face.git}}. The $l_2$-constrained Softmax loss was implemented follwing \cite{ranjan2017l2}  by integrating a normalization and scaling layer\footnote{see \url{https://github.com/craftGBD/caffe-GBD}. In our experiments, for $\alpha=50$ the gradients exploded due the relatively deep Res64 architecture and learning $\alpha$ initialized at 30 did not converge.} before the last fully-connected layer. For experiments with L-softmax \cite{liu2016large} and SphereFace \cite{liu2017sphereface}, we used the publicly available Caffe implementation. The Resnet 64 layer (Res64) architecture results in a feature dimension of 512 (at the fc5 layer), which is used for  matching using the cosine distance. Ring loss and Center loss are both applied on this feature \emph{i.e.} to the output of the fc5 layer.  All models were trained on the MS-Celeb 1M dataset \cite{guo2016ms}. The dataset was cleaned to remove potential outliers within each class and also noisy classes before training. To clean the dataset we used a pretrained model to  extract features from the MS-Celeb 1M dataset. Then, classes that had variance in the MSE, between the sample features and the mean feature of that class, above a certain threshold were discarded. Following this, from the filtered classes, images that have their MSE error between their feature vector and the class mean feature vector higher than a threshold are discarded. After this procedure, we are left with about 31,000 identities and about 3.5 million images. The learning rate was initialized to 0.1 and then decreased by a factor of 10 at 80K and 150K iterations for a total of 165K iterations. All models evaluated were the 165K iteration model\footnote{For all Tables, results reported after double horizontal lines are from models trained during our study. The results above the lines reported directly from the paper as cited.}.



\textbf{Preprocessing.} All faces were detected and aligned using \cite{zhang2016joint} which provided landmarks for the two eye centers, nose and mouth corners (5 points). Since MS-Celeb1M, IJB-A Janus and Janus CS3 have harder faces we use a robust detector \emph{i.e.} CMS-RCNN \cite{zhu2017cms} to detect faces and a fast landmarker that is robust to pose \cite{bhagavatula2017faster}. The faces were then aligned using a similarity transformation and were cropped to $112\times 96$ in the RGB format. The pixel level activations were normalized by subtracting 127.5 and then dividing by 128. For failed detections, the training set images are ignored. In the case of testing, ground truth landmarks were used from the corresponding dataset.



\begin{figure*}
    \begin{center}
        \subfigure[IJB-A Janus 1:1 Verification]{%
        \centering
            \includegraphics[width=0.9\columnwidth,valign=m]{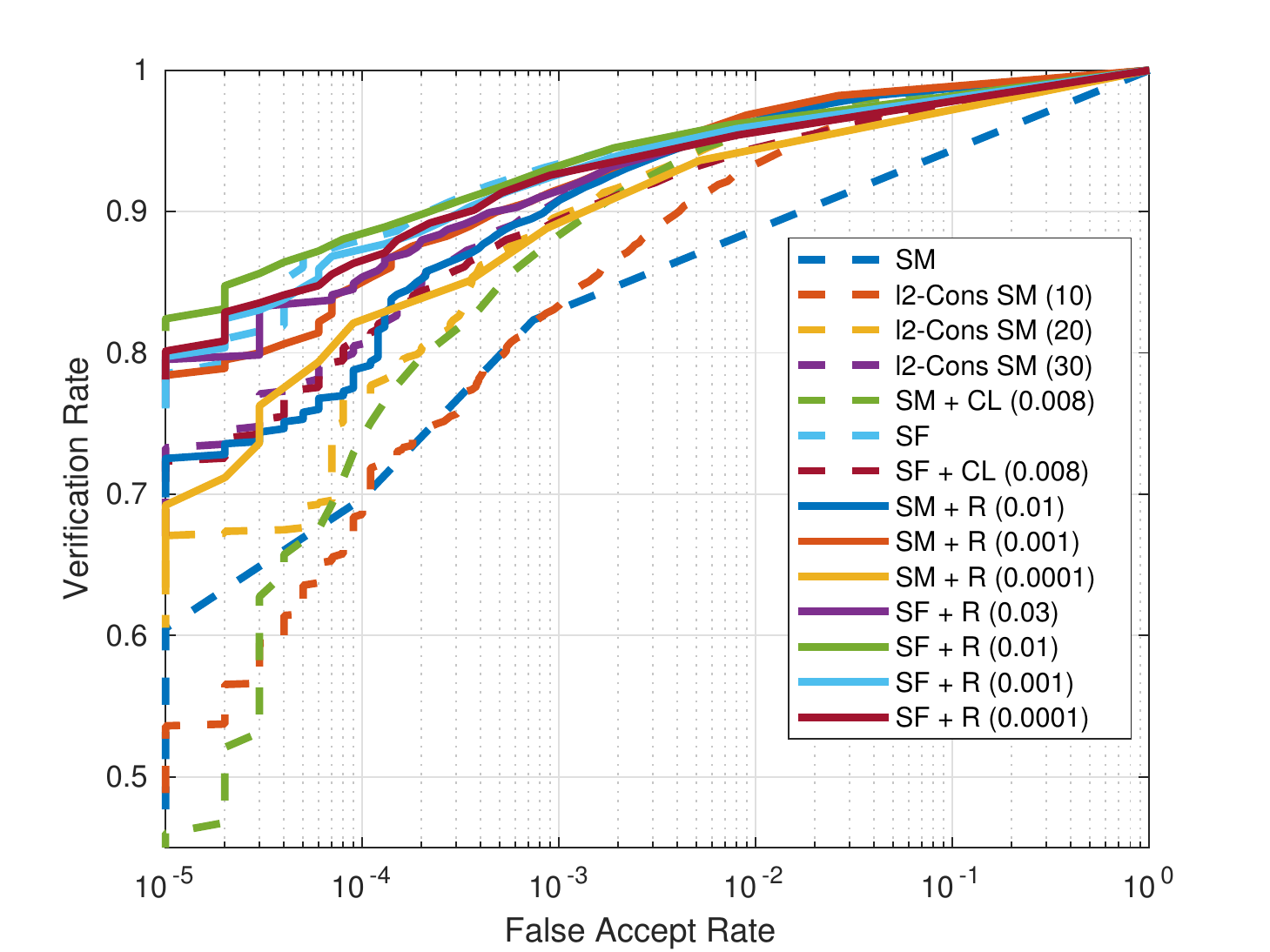}\label{fig_janus}
        }%
         \subfigure[Janus CS3 1:1 Verification]{%
        \centering
        \includegraphics[width=0.9\columnwidth,valign=m]{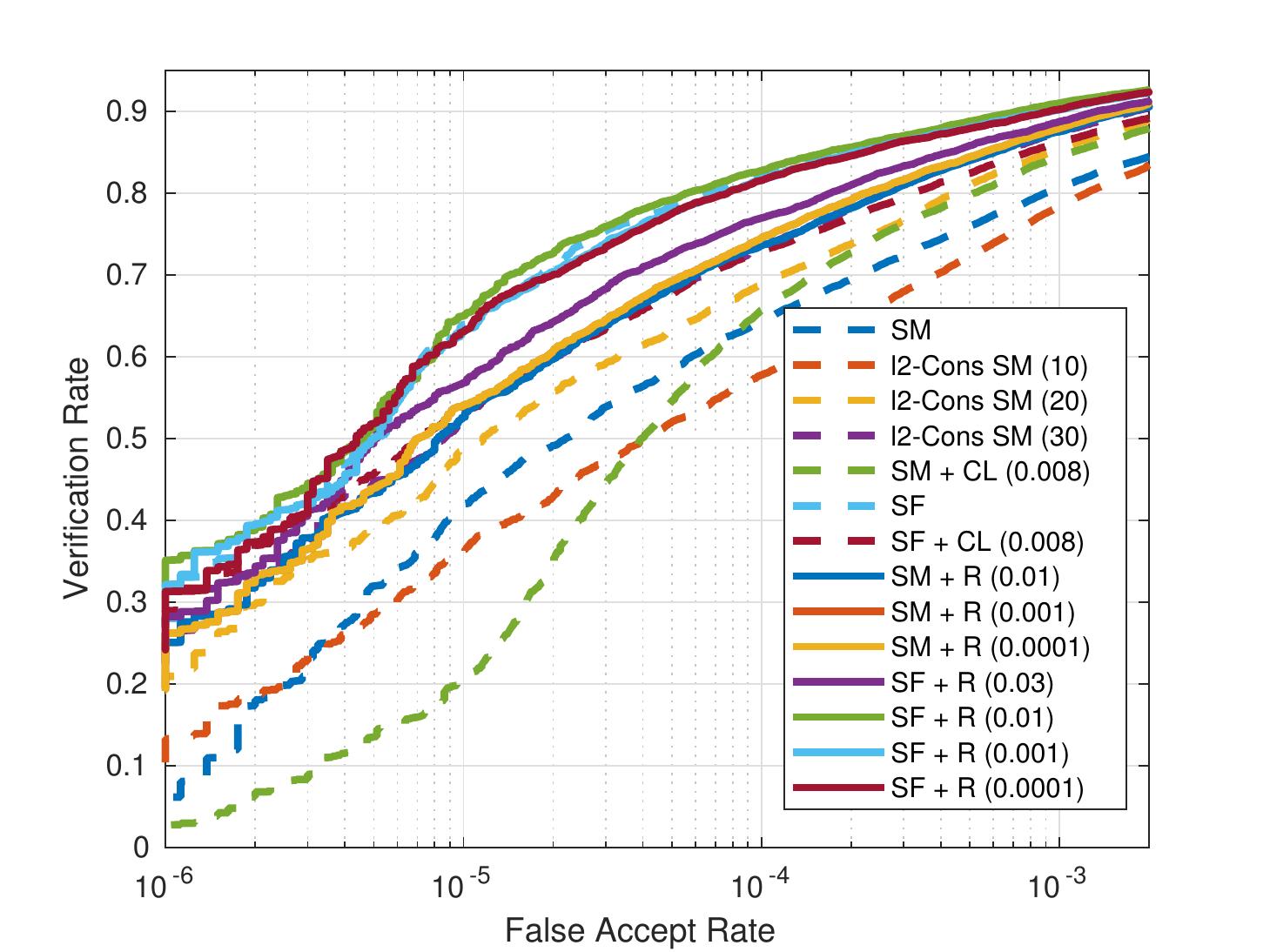}\label{fig_cs3}
        }
        
    \end{center}
    \vspace{-0.5cm}
\caption{ ROC curves on the (a) IJB-A Janus 1:1 verification protocol and the (b) Janus CS3 1:1 verification protocol. For all Figures and Tables, SM denotes Softmax, SF denotes SphereFace \cite{liu2017sphereface},l2-Cons SM denotes \cite{ranjan2017l2}, + CL denotes Center Loss augmentation \cite{wen2016discriminative} and finally + R denotes Ring loss augmentation. Numbers in bracket denote value of hyperparameter (loss weight), \emph{i.e.} $\alpha$ for \cite{ranjan2017l2}, $\lambda$ for Center loss and Ring loss.}
\vspace{-0.5cm}
\end{figure*}



\textbf{Exp 1. Testing Benchmark: LFW.} The LFW \cite{LFWTech} database contains about 13,000 images for about 1680 subjects with a total of 6,000 defined matches. The primary nuisance transformations are illumination, pose, color jittering and age. As the field has progressed, LFW has been considered to be saturated and prone to spurious minor variances in performance (in the last \% of accuracy) owing to the small size of the protocol. Small differences in accuracy on this protocol do not accurately reflect the generalizing capabilities of a high performing model. Nonetheless, as a benchmark, we report performance on this dataset.


\textbf{Results: LFW.}  Table~\ref{tab_lfw} showcases the results on the LFW protocol. We find that for Softmax (SM + R), Ring loss normalization seems to significantly improve performance (up from 98.47\% to \textbf{99.52\%} using Ring loss with $\lambda=0.01$). We find similar trends while using Ring loss with SphereFace. The LFW accuracy of SphereFace improves from 99.47\% to \textbf{99.50\%}. We note that since even our baselines are high performing, there is a lot of variance in the results owing to the small size of the LFW protocol (just 6000 matches compared to about 8 million matches in the Janus CS3 protocol which shows clearer trends). Indeed we find clearer trends with MegaFace, IJB-A and CS3 all of which are orders of magnitude larger protocols.

\textbf{Exp 2. Testing Benchmark: IJB-A Janus.} IJB-A \cite{klare2015pushing} is a challenging dataset which consists of 500 subjects with  extreme pose, expression and illumination with a total of 25,813 images. Each subject is described by a template instead of a single image. This allows for score fusion techniques to be developed. The setting is suited for applications which have multiple sources of images/video frames. We report results on the 1:1 template matching protocol containing 10 splits with about 12,000 pair-wise template matches each resulting in a total of 117,420 template matches. The template matching score for two templates $T_i, T_j$ is determined by using the following formula, $S(T_i, T_j) =  \sum_{\gamma=1}^K \frac{\sum_{t_a\in T_i, t_b\in T_j}  s(t_a, t_b)  \exp{ \gamma s(t_a, t_b) } }{  \sum_{t_a\in T_i, t_b\in T_j} \exp{ \gamma s(t_a, t_b) }   } $ where $s(t_a. t_b)$ is the cosine similarity score between images $t_a, t_b$ and $K=8$. 


\textbf{Results: IJB-A Janus.} Table.~\ref{tab_janus} and Fig.~\ref{fig_janus} present these results. We see that Softmax + Ring loss (0.001) outperforms Softmax by a large margin, particularly 60.52\% verification rate compared to \textbf{78.41\%} verification at $10^{-5}$ FAR. Further, it outperforms Center loss \cite{wen2016discriminative} (46.01\%) and $l_2$-constrained Softmax (73.29\%)  \cite{ranjan2017l2}. Although SphereFace performs better than Softmax + Ring loss, an augmentation by Ring loss boosts SphereFace's performance from 78.52\% to \textbf{82.41\%} verification rate for $\lambda=0.01$. This matches the state-of-the-art reported in \cite{ranjan2017l2} which uses a 101-layer ResNext architecture despite our system using a much shallower 64-layer ResNet architecture. The effect of high $\lambda$ akin to the effects simulated in Fig.~\ref{fig_ring_loss_visual} show in this setting for $\lambda=0.03$ for SphereFace augmentation. We observe this trade-off in Janus CS3, CFP and MegaFace results as well. Nonetheless, we notice that Ring loss augmentation provides consistent improvements over a large range of $\lambda$ for both Softmax and Sphereface. This is in sharp contrast with  $l_2$-constrained Softmax whose performance varies significantly with $\alpha$ rendering it difficult to optimize. In fact for $\alpha=10$, it performs worse than Softmax. 



\textbf{Exp 3. Testing Benchmark: Janus CS3.} The Janus CS3 dataset is a super set of the IARPA IJB-A dataset. It contains about 11,876 still images and 55,372 video frames from 7,094 videos. For the CS3 1:1 template verification protocol there are a total of 1,871 subjects and 12,590 templates. The CS3 template verification protocol has over 8 million template matches which amounts to an extremely large number of template verifications. There are about 1,870 templates in the gallery and about 10,700 templates in the probe. The CS3 protocol being a super set of IJB-A, has a large number of extremely challenging images and faces. The challenging conditions range from extreme illumination, extreme pose to significant occlusion. For sample images, please refer to Fig. 4 and Fig. 10 in \cite{lin2017proximity}, Fig. 1 and Fig. 6 in \cite{bodla2017deep}. Since the protocol is template matching, we utilize the same template score fusion technique we utilize in the IJB-A results with $K=2$.

\begin{figure*}
    \begin{center}
        \subfigure[Downsampling 4x]{%
        \centering
            \includegraphics[width=0.4\columnwidth,valign=m]{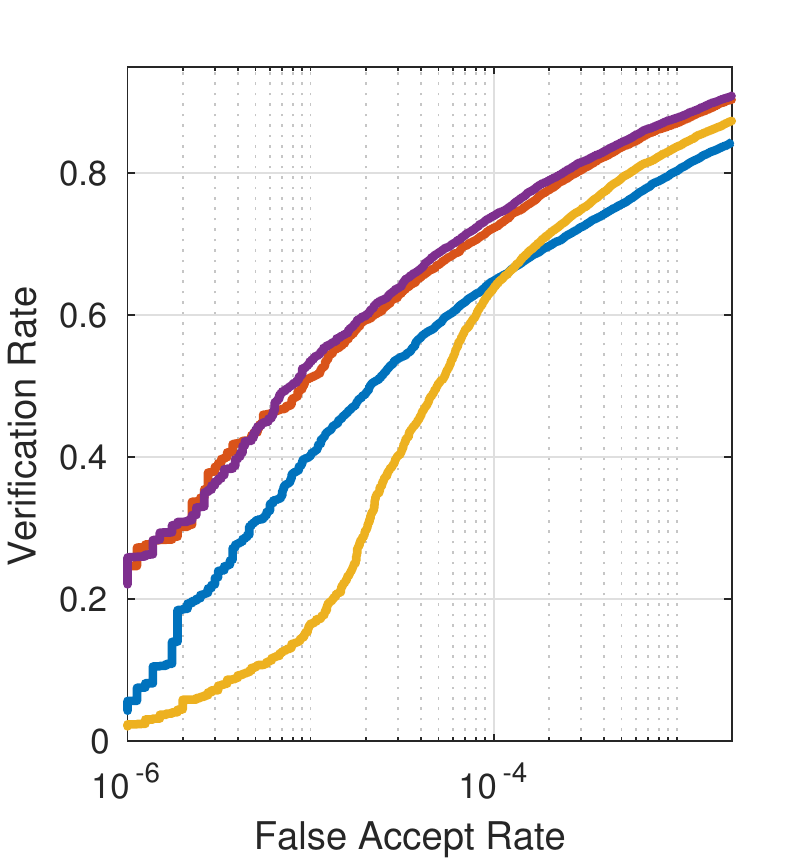}
        }%
         \subfigure[Downsampling 16x]{%
        \centering
        \includegraphics[width=0.4\columnwidth,valign=m]{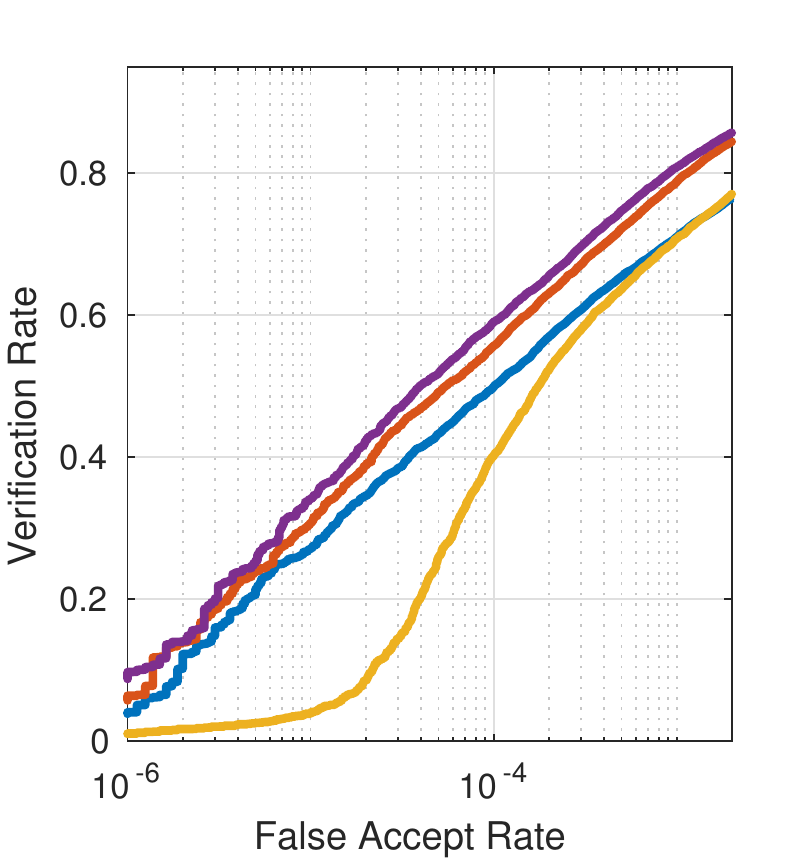}
        }
         \subfigure[Downsampling 25x]{%
        \centering
        \includegraphics[width=0.4\columnwidth,valign=m]{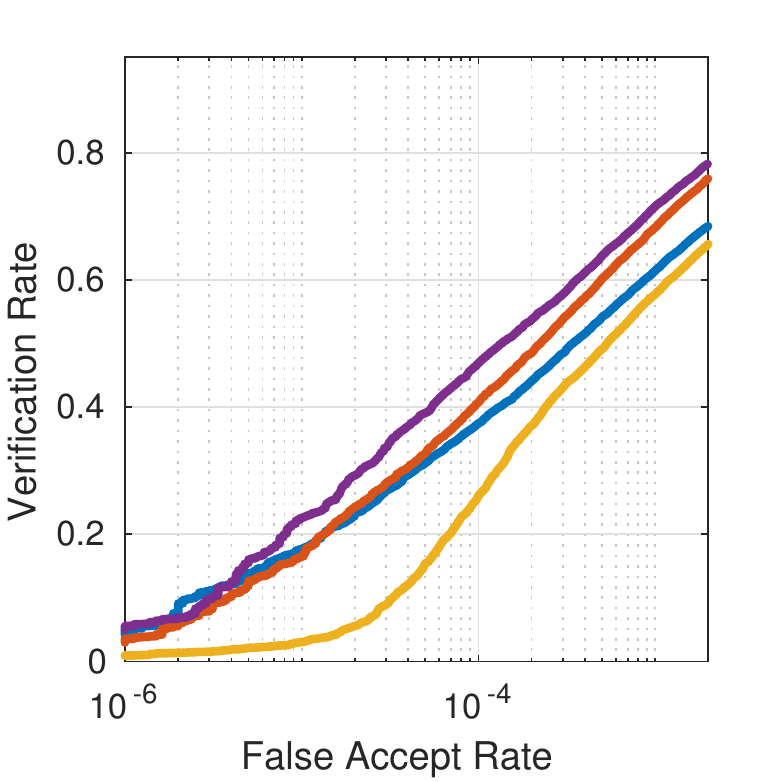}
        }
         \subfigure[Downsampling 36x]{%
        \centering
        \includegraphics[width=0.4\columnwidth,valign=m]{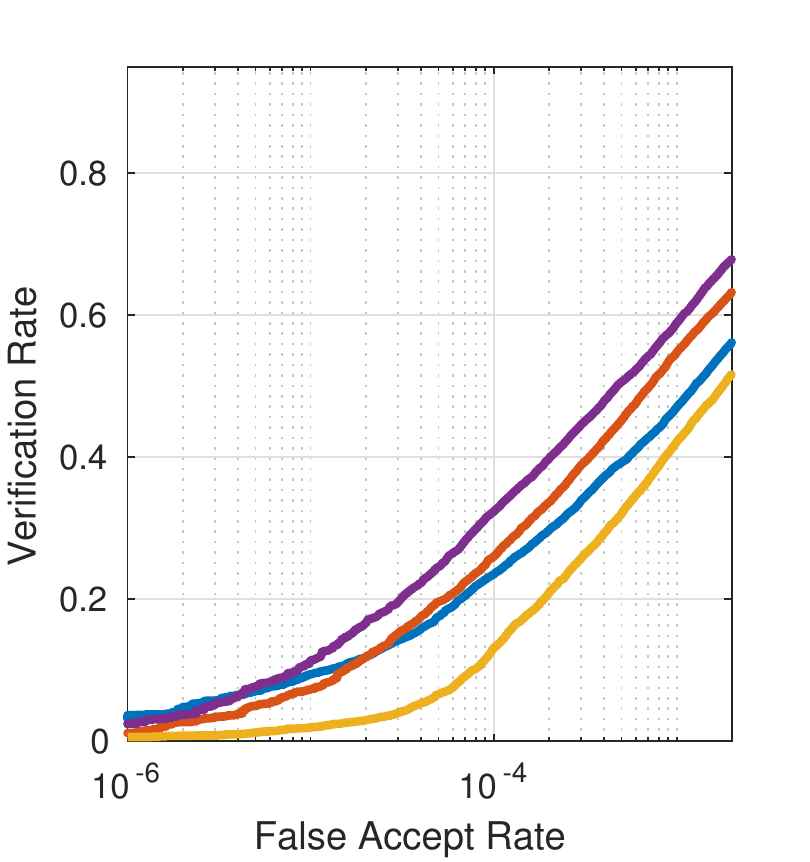}
        }
         \subfigure[Downsampling 64x]{%
        \centering
        \includegraphics[width=0.4\columnwidth,valign=m]{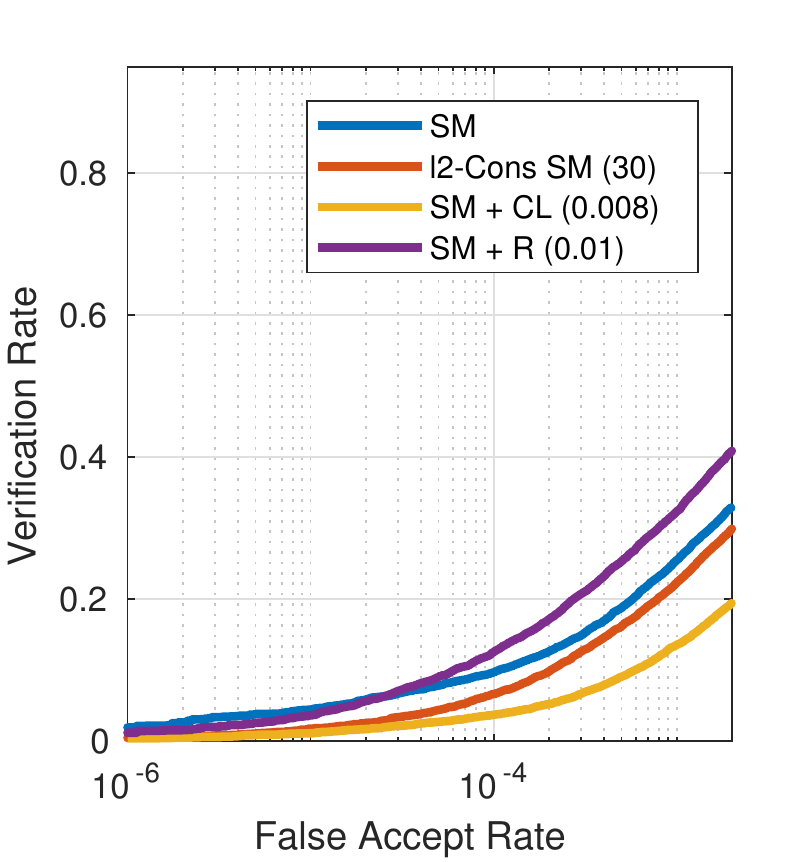}
        }
    \end{center}
    \vspace{-0.5cm}
\caption{ ROC curves for the downsampling experiment on Janus CS3. Ring loss (SM + R $\lambda=0.01$) learns the most robust features, whereas $l_2$-constrained Softmax (l2-Cons SM $\alpha=30$) \cite{ranjan2017l2} performs poorly (worse than the baseline Softmax) at very high downsampling factor of 64x. }
\label{fig_scale2}
\vspace{-0.5cm}
\end{figure*}

\textbf{Results: Janus CS3.} Table.~\ref{tab_cs3} and Fig.~\ref{fig_cs3} showcases our results on the CS3 dataset. We report verification rates (VR) at $10^{-3}$ through $10^{-6}$ FAR. We find that our Ring loss augmented Softmax model outperforms the previous best reported results on the CS3 dataset. Recall that the Softmax + Ring loss model (SM + R) was trained only on a subset of the MS-Celeb dataset and achieves a VR of 74.56\% at $10^{-4}$ FAR. This is in contrast to Lin \emph{et. al.} who train on MS-Celeb plus CASIA-WebFace (an additional 0.5 million images) and achieve 72.52 \%. Further, we find that even though our baseline Sphereface Res64 model outperforms the previous state-of-the-art, our Ring loss augmented Sphereface model outperforms all other models to achieve high a VR of \textbf{82.74 \%} at $10^{-4}$ FAR. At very low FAR of $10^{-6}$ our SF + R model achieves VR \textbf{35.18 \%} which to the best of our knowledge is the state-of-the-art on the challenging Janus CS3. In accordance with the results on IJB-A Janus, Ring loss provides consistent improvements over large ranges of $\lambda$ whereas $l_2$-constrained Softmax exhibits significant variation w.r.t. to its hyperparameter.

\textbf{Exp 4. Testing Benchmark: MegaFace. } The recently released MegaFace benchmark is extremely challenging which defines matching with a gallery of about 1 million distractors \cite{kemelmacher2016megaface}. The aspect of face recognition that this database test is discrimination in the presence of very large number of distractors. The testing database contains two sets of face images. The distractor set contains 1 million distractor subjects (images). The target set contain 100K images from 530 celebrities.


\textbf{Result: MegaFace.}  Table.~\ref{tab_megaface} showcases our results. Even at this extremely large scale evaluation (evaluating FaceScrub against 1 million), the addition of Ring loss provides significant improvement to the baseline approaches. The identification rate ($\%$) for Softmax upon the addition of Ring loss ($\lambda = 0.001$) improves from 56.36\% to a high \textbf{71.67\%} and for SphereFace it improves from 74.95\% to \textbf{75.22\%} for a \textit{single} patch model. This is higher than the single patch model reported in the orginal Sphereface paper (72.72\% \cite{liu2017sphereface}). We outperform Center loss  \cite{wen2016discriminative} augmenting both Softmax (67.24\%) and Sphereface (71.15\%). We find that though for MegaFace, $l_2$-constrained Softmax \cite{ranjan2017l2} for $\alpha=30$  achieves 72.22\%, there is yet again significant variation in performance that occurs due to a change in the hyper parameter $\alpha$ (66.20\% for $\alpha=10$ to 72.22\% for $\alpha=30$). Ring loss hyper parameter ($\lambda$), as we find again, is more easily tunable and manageable. This results in a smaller variance in performance for both Softmax and SphereFace augmentations.


\begin{table}
\small
\caption{Accuracy (\%) on LFW. } 
\centering 
\begin{tabular}{l c c } 

\hline\hline 

Method & Training Data   & Accuracy (\%)   \\
\hline 
FaceNet \cite{schroff2015facenet}  & 200M private  & 99.65   \\ 
Deep-ID2+  \cite{sun2014deep}  &  CelebFace+	& 99.15  \\
Range loss \cite{ZhangFWL016}  &  WebFace &  99.52 \\
  &  +Celeb1M(1.5M) &  \\
Baidu \cite{liu2015targeting}  &  1.3M &  99.77 \\

Norm Face \cite{Wang2017NormFace} & WebFace  &   99.19 \\

\hline
\hline

SM    &   MS-Celeb     &   98.47  \\
$l2$-Cons SM (30) \cite{ranjan2017l2}   &   MS-Celeb     &   99.55  \\ 
$l2$-Cons SM (20) \cite{ranjan2017l2}   &   MS-Celeb     &   99.47  \\ 
$l2$-Cons SM (10) \cite{ranjan2017l2}   &   MS-Celeb     &   99.45  \\ 
SM + CL \cite{wen2016discriminative}   &  MS-Celeb     &     99.17   \\ 
SF   \cite{liu2017sphereface}   &   MS-Celeb     &   99.47   \\
SF + CL \cite{wen2016discriminative, liu2017sphereface}  &  MS-Celeb     &   99.52   \\

\hline
SM + R  ($0.01$)    &  MS-Celeb     & 99.52    \\
SM + R  ($0.001$)   &  MS-Celeb     &  99.50  \\
SM + R  ($0.0001$)  & MS-Celeb   &    99.28    \\ 
\hline
SF + R  ($0.03$)   &   MS-Celeb     &    99.48  \\ 
SF + R  ($0.01$)   &   MS-Celeb     &    99.43 \\ 
SF + R  ($0.001$)   &   MS-Celeb     &   99.42 \\ 
SF + R  ($0.0001$)   &  MS-Celeb     &  99.50   \\ 

\hline
\end{tabular}
\label{tab_lfw} 
\end{table}
\begin{table}
\centering
\begin{tabular}{l  c c } 

\hline\hline 
Method   & Acc \% (MegaFace)  &  $10^{-3}$ (CFP)  \\
\hline
\hline

SM    &     56.36  &   55.86   \\
$l2$-Cons SM (30) \cite{ranjan2017l2}   &  72.22 &   82.14   \\ 
$l2$-Cons SM (20) \cite{ranjan2017l2}   &   70.29    &  83.69  \\ 
$l2$-Cons SM (10) \cite{ranjan2017l2}   &   66.20   &   76.77  \\ 
SM + CL \cite{wen2016discriminative}   &  67.24     & 78.94   \\ 
SF   \cite{liu2017sphereface}   &   74.95  &   89.94  \\
SF + CL \cite{wen2016discriminative, liu2017sphereface}  &  71.15  &   82.97 \\

\hline
SM + R  ($0.01$)    &  71.10  &  87.43  \\
SM + R  ($0.001$)   &  71.67 &  81.29  \\
SM + R  ($0.0001$)  &   69.41  &  76.30  \\ 
\hline
SF + R  ($0.03$)   &    73.05 & 86.23    \\ 
SF + R  ($0.01$)   &     74.93  & \textbf{ 90.94}   \\ 
SF + R  ($0.001$)   &   \textbf{ 75.22 } &  87.69   \\ 
SF + R  ($0.0001$)   &   74.45  &   88.17 \\

\hline
\end{tabular}
\caption{Identification rates on MegaFace with 1 million distractors (Accuracy \%) and Verification rates at $10^{-3}$ FAR for the CFP Frontal vs. Profile protocol.  }
\label{tab_megaface}

\end{table}

\begin{table}

\hfill
\centering
\begin{tabular}{l c c c} 

\hline\hline 

Method  & $10^{-5}$   & $10^{-4}$  &  $10^{-3}$  \\
\hline 

$l2$-Cons SM* (101) \cite{ranjan2017l2} &-     &    87.9  &  93.7  \\ 
$l2$-Cons SM* (101x) \cite{ranjan2017l2} & -& 88.3     & 93.8  \\

\hline 
\hline





SM    &    60.52   &  69.69     & 83.10   \\ 

$l2$-Cons SM (30)  \cite{ranjan2017l2}&  73.29   & 80.65   &   90.72   \\ 
$l2$-Cons SM (20)  \cite{ranjan2017l2}&   67.63  &   76.88    & 89.89     \\ 
$l2$-Cons SM (10)  \cite{ranjan2017l2}&  53.74   & 68.58    &  83.42    \\ 
SM + CL \cite{wen2016discriminative} &  46.01  & 74.10  &   88.32  \\ 
SF   \cite{liu2017sphereface}& 78.52    &  88.0   &  \textbf{93.24}  \\
SF + CL \cite{wen2016discriminative, liu2017sphereface}  & 72.35   & 81.11  &  89.26 \\
\hline

SM + R ($0.01$) & 72.53  &  79.1  & 90.8     \\ 
SM + R ($0.001$) &   78.41  &  85.0  &  91.5      \\ 
SM + R  ($0.0001$)  &  69.23  &  82.30   &  89.20    \\ 

\hline
SF + R  ($0.03$) & 79.54   &    85.37  &  91.64    \\ 
SF + R  ($0.01$) &   \textbf{82.41}   & \textbf{ 88.5} & \textbf{93.22}     \\ 

SF + R  ($0.001$) & 79.74    &     87.71 &  92.62    \\
SF + R  ($0.0001$) &    80.13 &     86.34    & 92.57    \\





\hline
\end{tabular}
\caption{Verification \% on the IJB-A Janus 1:1 verification protocol.  $l2$-Cons SM* indicates the result reported in \cite{ranjan2017l2} which uses a 101 layer ResNet/ResNext architecture.}
\label{tab_janus}
\end{table}


\begin{table}
\hfill
\centering
\begin{tabular}{l c c c c} 

\hline\hline 

Method  & $10^{-6}$  &  $10^{-5}$    &  $10^{-4} $   &  $10^{-3}$    \\
\hline 

Bodla \emph{et. al.} Final1 \cite{bodla2017deep}   &  -  & -   &  69.81  & 82.89    \\
Bodla \emph{et. al.} Final2 \cite{bodla2017deep}   &  -   &  -   &  68.45  & 82.97   \\

Lin \emph{et. al.}  \cite{lin2017proximity}    &  -  &  -  & 72.52   &  83.55 \\

\hline
\hline
SM  &  6.16   & 42.03    & 64.52    &  80.86   \\ 
$l2$-Cons SM (30)  \cite{ranjan2017l2}&  24.47  & 52.32   & 73.36    &  87.46  \\ 
$l2$-Cons SM (20)  \cite{ranjan2017l2}&  21.14  & 48.82  &   68.84  &  85.34    \\ 
$l2$-Cons SM (10)  \cite{ranjan2017l2}& 13.28  & 36.08  & 57.80  & 78.36   \\ 
SM + CL \cite{wen2016discriminative} &  2.88  & 20.87   &  65.71    &  84.55   \\ 
SF  \cite{liu2017sphereface} &  28.51   &  63.92 &   82.29  &  90.58  \\
SF + CL \cite{wen2016discriminative, liu2017sphereface}  &  28.99  &    53.36    &  72.91  &    86.14 \\

\hline

SM + R ($0.01$)   & 25.17  & 52.60  & 73.56  &  87.50  \\
SM + R  ($0.001$)  &  26.62 &  54.13    & 74.56    &  87.93     \\ 
SM + R  ($0.0001$)  & 17.35 &  50.65   &  71.06  &  85.48   \\ 
\hline
SF + R  ($0.03$)  &   27.27  &  56.84    &  76.97   &  88.75   \\ 
SF + R  ($0.01$)  & \textbf{ 35.18 } &  \textbf{65.02}  &\textbf{ 82.74 } &   \textbf{90.99 }    \\ 
SF + R  ($0.001$)  & 32.19    &  63.13     &    81.62   &  90.17   \\
SF + R  ($0.0001$)  &   32.01  & 63.12    & 81.57   &  90.24  \\ 




\hline
\end{tabular}
\caption{Verification \% on the Janus CS3 1:1 verification protocol. }
\label{tab_cs3}
\end{table}


\textbf{Exp 5. Testing Benchmark: CFP Frontal vs. Profile. } Recently the CFP (Celebrities Frontal-Profile) dataset was released to evaluate algorithms exclusively on frontal versus profile matches \cite{sengupta2016frontal}. This small dataset has about 7,000 pairs of matches defined with 3,500 same pairs and 3,500 not-same pairs for about 500 different subjects. For sample images please refer to Fig. 1 of \cite{sengupta2016frontal}. The dataset presents a challenge since each of the probe images is almost entirely profile thereby presenting extreme pose along with illumination and expression challenges. 


\textbf{Result: CFP Frontal vs. Profile.} Fig.~\ref{fig_cfp} showcases the ROC curves for this experiment whereas Table.~\ref{tab_megaface} shows the verification rates at $10^{-3}$ FAR.  Ring loss (\textbf{87.43\%}) provides consistent and significant boost in performance over Softmax (55.86\%). We find however, SphereFace required more careful tuning of $\lambda$ with $\lambda=0.01$ (\textbf{90.94\%}) outperforming the baseline. Further, Softmax and Ring loss with $\lambda=0.01$ significantly outperforms all runs for $l_2$-constrained Softmax \cite{ranjan2017l2} (83.69). Thus, Ring loss helps in providing higher verification rates while dealing with frontal to highly off-angle matches thereby explicitly demonstrating robustness to pose variation.

\textbf{Exp 6. Low Resolution Experiments on Janus CS3. } One of the main motivations for $l_2$-constrained Softmax was to handle images with varying resolution. Low resolution images were found to result in low norm features and vice versa. Ranjan \emph{et.al.}  \cite{ranjan2017l2} argued normalization (through $l_2$-constrained Softmax) would help deal with this issue. In order to test the efficacy of our alternate convex normalization formulation towards handling low resolution faces, we synthetically downsample Janus CS3 from an original size of ($112\times 96$) by a factor of 4x, 16x, 25x, 36x and 64x respectively (images were downsampled and resized back up using bicubic interpolation in order to fit the model). We run the Janus CS3 protocol and plot the ROC curves in Fig.~\ref{fig_scale2}. We find that the Ring loss helps Softmax features be more robust to resolution. Though $l_2$-constrained Softmax provides improvement over Softmax, it's performance is lower than Ring loss. Further, at extremely high downsampling of 64x, $l_2$-constrained Softmax in fact performs worse than Softmax, whereas Ring loss provides a clear improvement. Center loss fails early on at 16x. We therefore find that our simple convex soft normalization approach is more effective at arresting performance drop due to resolution in accordance with the motivation for as normalization presented in \cite{ranjan2017l2}.

\textbf{Conclusion.} We motivate feature normalization in a principled manner and develop an elegant, simple and straight forward to implement convex approach towards that goal. We find that Ring loss consistently provides significant improvements over a large range of the hyperparameter $\lambda$. Further, it helps the network itself to learn normalization thereby being robust to a large range of degradations.

{\small
\bibliographystyle{ieee}
\bibliography{egbib}
}

\end{document}